\pdfoutput=1
\documentclass[11pt]{article}

\usepackage{emnlp2021}

\usepackage{times}
\usepackage{latexsym}

\usepackage[T1]{fontenc}
\usepackage[utf8]{inputenc}

\usepackage{microtype}

\usepackage{graphicx}
\usepackage{latexsym}
\usepackage{amsmath,amssymb} % define this before the line numbering.
\usepackage{xcolor}
\usepackage{booktabs}
\usepackage{subcaption}
\usepackage{multirow}
\usepackage{subcaption}
\usepackage{arydshln}

\title{CLIP4Clip: An Empirical Study of CLIP for End to End Video Clip Retrieval}
\author{
	Huaishao Luo\textsuperscript{1}\thanks{~This work was done during the first author's internship in MSR Asia}~,
	Lei Ji\textsuperscript{2},
	Ming Zhong\textsuperscript{3},
	Yang Chen\textsuperscript{3},
	Wen Lei\textsuperscript{3},
	Nan Duan\textsuperscript{2},
	Tianrui Li\textsuperscript{1} \\
	\textsuperscript{1}Southwest Jiaotong University, Chengdu, China\\
	{\tt huaishaoluo@gmail.com, trli@swjtu.edu.cn} \\
	\textsuperscript{2}Microsoft Research Asia, Beijing, China \\ 
	\textsuperscript{3}Microsoft STCA, Beijing, China \\ 
	{\tt \{leiji,minzhon,emchen,wen.lei,nanduan\}@microsoft.com}
}

\begin{document}
	\maketitle
	\begin{abstract}
		Video-text retrieval plays an essential role in multi-modal research and has been widely used in many real-world web applications.  The CLIP (Contrastive Language-Image Pre-training), an image-language pre-training model, has demonstrated the power of visual concepts learning from web collected image-text datasets. In this paper, we propose a CLIP4Clip model to transfer the knowledge of the CLIP model to video-language retrieval in an end-to-end manner. Several questions are investigated via empirical studies: 1) Whether image feature is enough for video-text retrieval? 2) How a post-pretraining on a large-scale video-text dataset based on the CLIP affect the performance? 3) What is the practical mechanism to model temporal dependency between video frames? And 4) The Hyper-parameters sensitivity of the model on video-text retrieval task. Extensive experimental results present that the CLIP4Clip model transferred from the CLIP can achieve SOTA results on various video-text retrieval datasets, including MSR-VTT, MSVC, LSMDC, ActivityNet, and DiDeMo. We release our code at \url{https://github.com/ArrowLuo/CLIP4Clip}.
	\end{abstract}

	\section{Introduction}
	With the increasing of videos uploaded online every day, video-text retrieval is becoming an emerging requirement for people to find relevant videos efficiently. Beyond the actual web application, video-text retrieval is a fundamental research task for multi-modal visual and language understanding. We can distinguish the previous works directly by their input: raw video (pixel-level) or video feature (feature-level).

	Usually, the pretrain models \cite{Zhu_2020_CVPR,Luo2020UniVL,Li2020HERO,Gabeur2020MMT,patrick2021supportset,Rouditchenko2020} are feature-level because they are trained on some large-scale video-text datasets, e.g., Howto100M \cite{miech2019howto100m}. The input is the cached video features generated via off-the-shelf frozen video feature extractors. If the input is the raw video, it makes the pretrain very slow or infeasible. Nevertheless, benefitting from the large-scale dataset, the pretrain models show a significant performance gain for video-text retrieval.

	The pixel-level approach trains the model with raw video as input directly \cite{Torabi2016Learning,kiros2014unifying,Yu2016Video,Kaufman2017Temporal,Yu2017End,yu2018joint}. Early literature almost belongs to this approach. This approach learns the video feature extractor jointly with the paired text. On the contrary, the feature-level approach highly depends on a suitable feature extractor. It can not propagate the learning back to the fixed video encoder.

	Some recent works begin to pretrain the model with the pixel-level approach, making the pretrain model learn from the raw video. The big challenge is how to reduce the high computational overload of dense video input. Typically, ClipBERT \cite{lei2021less} employs a sparse sampling strategy to make the end-to-end pretrain possible. Concretely, the model only sparsely samples a single or a few short clips from a video at each training step. The performance shows that the end-to-end training can benefit the low-level feature extraction. A few sparsely sampled clips are enough to solve the video-text retrieval task. Frozen \cite{Bain2021Frozen} treats an image as a single-frame video and designs a curriculum learning schedule to train the model on both image and video datasets. The results show that the curriculum learning schedule learns increasingly from image to multi frames can increase efficiency. Our target is not to pretrain a new model on video-text retrieval. We mainly investigate how to transfer the knowledge from the image-text pretrained model CLIP \cite{radford2021learning} to video-text retrieval in this paper.

	We exploit the pre-trained CLIP and propose a model named \textbf{CLIP4Clip} (\textbf{CLIP} \textbf{For} video \textbf{Clip} retrieval) to solve video-text retrieval. Concretely, the CLIP4Clip is constructed on top of the CLIP and designs a similarity calculator to investigate three similarity calculation approaches: parameter-free type, sequential type, and tight type. Like our work, the concurrent work from \citet{PortilloQuintero2021} is also built on the CLIP for video-text retrieval. The difference is that their work directly leveraged the CLIP for zero-shot prediction without considering different similarity calculation mechanisms. However, we design some similarity calculation approaches to improve the performance and train the model in an end-to-end manner. The contributions of our work are: 1) we investigate three mechanisms of similarity calculation based on the pretrained CLIP; 2) we further post pre-train the CLIP on a noisy large-scale video-language dataset to learn a better retrieval space. The extensive experiments show our model achieves the new SOTA results on MSR-VTT \cite{xu2016msr}, MSVC \cite{chen2011collecting}, LSMDC \cite{Rohrbach2015LSMDC}, ActivityNet \cite{krishna2017dense}, and DiDeMo \cite{hendricks17iccv} datasets.

	Besides, we can conclude the following insights from our extensive experiments:

	1)  One single image is far from enough for video encoding for video-text retrieval.

	2)  Post-pretraining at a large-scale video-text dataset on the CLIP4Clip model is required and can improve the performance, especially for zero-shot prediction by a large margin.

	3)  With the powerful pre-trained CLIP, it is better not to introduce new parameters and adopt a mean pooling mechanism on video frames for small datasets. At the same time, it is better to introduce more parameters, e.g., the self-attention layer, to learn the temporal dependency for large datasets.

	4)  We carefully study the hyper-parameters and report the best setting.

	\section{Related Works}
	\paragraph{Video Encoder Backbone}
	Previous works mainly focus on 2D/3D spatial-temporal convolution for video representation \cite{tran2015learning,Xie2018Rethinking,feichtenhofer2019slowfast}. Recently ViT \cite{dosovitskiy2021an}, a transformer-based image encoder, has attracted much attention. The transformer based video encoder is still in its early stage for action classification \cite{bertasius2021spacetime, arnab2021vivit}. We mainly investigate the effective transformer based video backbone for multimodal video-text retrieval.
	\begin{figure*}[htbp]
		\centering
		\subfloat[Main structure] {
			\centering
			\includegraphics[width=0.99\textwidth]{figures/Framework7.pdf}
			\label{fig:structure}
		} \\
		\subfloat[Similarity calculator] {
			\centering
			\includegraphics[width=0.88\textwidth]{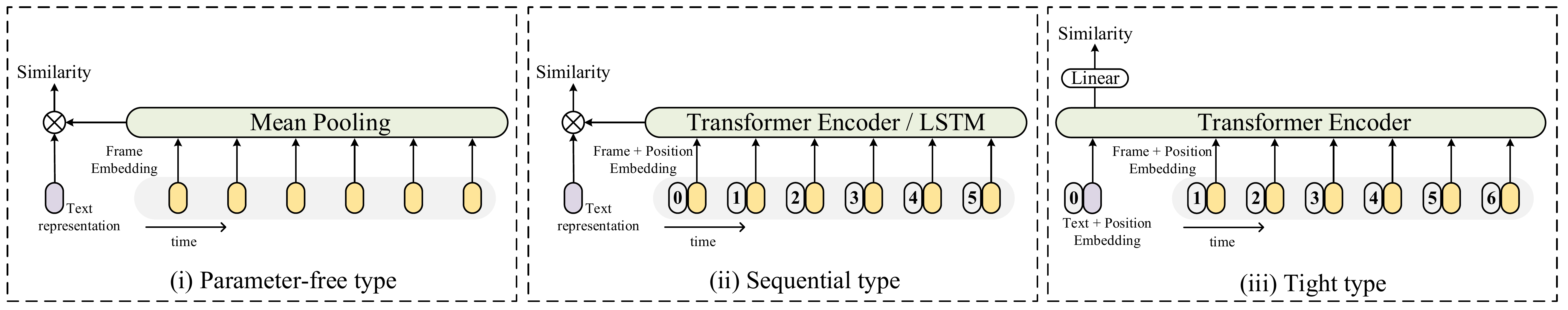}
			\label{fig:similarity_calculator}
		}
		\caption{The framework of CLIP4Clip, which comprises three components, including two single-modal encoders and a similarity calculator. The model takes a video-text pair as input. For the input video, we first sample the input video into ordinal frames (images). Next, these image frames are reshaped into a sequence of flattened 2D patches. These patches are mapped to the 1D sequence of embeddings with a linear patch embedding layer and input to the image encoder for representation as in ViT \cite{dosovitskiy2021an}. Finally, the similarity calculator predicts the similarity score between the text representation and representation sequence of these frames. We investigate three types of similarity calculators in this work, including parameter-free, sequential, and tight types. $\otimes$ means cosine similarity. We initial the two single-modal encoders with CLIP (ViT-B/32) \cite{radford2021learning}.}
		\label{fig:main_structure}
	\end{figure*}

	\paragraph{Visual Representation Learning from Text Supervision}
    Visual representation learning is a challenging task and has been widely studied with supervised or self-supervised methods. Considering semantic supervision from large-scale unlabeled data, learning visual representation from text representation \cite{radford2021learning, miech19endtoend, lei2021less} is an emerging research topic with the benefit of large-scale visual and linguistic pairs collected from the Internet. The prominent success of the CLIP (Contrastive Language-Image Pre-training) \cite{radford2021learning} has demonstrated its capability of learning SOTA image representations from linguistic supervision with pre-training on large-scale image and text pairs. The pre-trained model can learn fine-grained visual concepts for images and transfer the knowledge for the retrieval task. Typically, MIL-NCE \cite{miech19endtoend} mainly investigated to leverage noisy large-scale Howto100M \cite{miech2019howto100m} instructional videos to learn a better video encoder in an end-to-end manner. Furthermore, ClipBERT \cite{lei2021less} proposed an efficient end-to-end approach through sparse sampling and presented that the pretrained by image-language dataset facilitated a better initialization of video-text retrieval. Different from ClipBERT, we adopt CLIP with transformer based visual backbone and extend this image-language pre-trained model to video-language pre-training for video-text retrieval. Considering the temporal sequence of video, a 2D/3D linear embedding and a similarity calculator attached to the visual transformer are used to capture temporal sequence features.

    \paragraph{Video-Text Retrieval}
	Early works on video-text retrieval \cite{Torabi2016Learning,kiros2014unifying,Yu2016Video,Kaufman2017Temporal,Yu2017End,yu2018joint}designed intensive fusion mechanisms for cross-modal learning. Recently, the pre-trained models \cite{Zhu_2020_CVPR,Amrani2020Noise,Luo2020UniVL,Li2020HERO,miech19endtoend,Gabeur2020MMT,patrick2021supportset,lei2021less,Dzabraev2021MDMMT,Liu2021HiT} have dominated the leaderboard of the video-text retrieval with noticeable results on zero-shot retrieval and fine-tuned retrieval. Concurrent to our work, \citet{PortilloQuintero2021} applied CLIP for zero-shot prediction, and \citet{Bain2021Frozen} proposed a transformer-based video backbone. We propose to directly transfer the powerful knowledge from the pre-trained CLIP and continue pre-train the designed video-based CLIP4Clip on a large-scale video-language dataset. Empirical studies present the effectiveness of the CLIP4Clip model.

	\section{Framework}
	Given a set of videos (or video clips) $\mathcal{V}$ and a set of captions $\mathcal{T}$, our target is to learn a function $s(v_i, t_j)$ to calculate the similarity between the video (or video clip) $v_i \in \mathcal{V}$ and the caption $t_j \in \mathcal{T}$. The goal is to rank all the videos (or video clips) given the query caption according to their similarity score in the text-to-video retrieval or rank all the captions given the query video (or video clip) in the task of video-to-text retrieval. The objective of the $s(v_i, t_j)$ is to calculate a high similarity for relevant video-text pairs and a low similarity score for irrelevant ones.

	The video (or video clip) $v_i \in \mathcal{V}$ is represented as a sequence of frames (images) in this paper. Formally, the video (or video clip) $v_i$ is composed of $|v_i|$ sampled frames such that $v_i=\{v^1_i, v^2_i, \dots, v^{|v_i|}_i\}$. Our model is an end-to-end manner (E2E) trained on pixels directly via taking the frames as input. Figure \ref{fig:main_structure} demonstrates our framework, which mainly contains a text encoder, a video encoder, and a similarity calculator. We introduce each part in detail in this section.

	\subsection{Video Encoder}
	\label{sec:video_encoder}
	To get the video representation, we first extract the frames from the video clip and then encode them via a video encoder to obtain a sequence of features. In this paper, we adopt the ViT-B/32 \cite{dosovitskiy2021an} with 12 layers and the patch size 32 as our video encoder. Concretely, we use the pre-trained CLIP (ViT-B/32) \cite{radford2021learning} as our backbone and mainly consider transferring the image representation to video representation. The pre-trained CLIP (ViT-B/32) is effective for the video-text retrieval task in this paper.
	\begin{figure}[tbp]
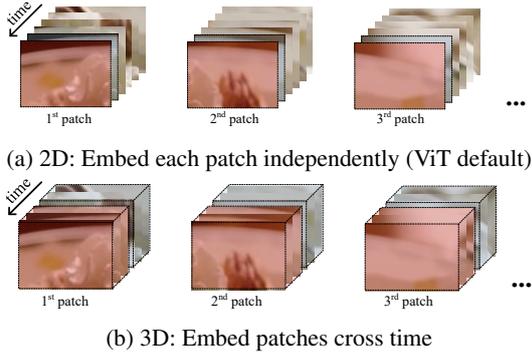

		\centering
		\subfloat[2D: Embed each patch independently (ViT default)] {
			\centering
			\includegraphics[width=0.44\textwidth]{figures/linear_2d.pdf}
			\label{fig:fig_linear_2d}
		} \\
		\subfloat[3D: Embed patches cross time] {
			\centering
			\includegraphics[width=0.44\textwidth]{figures/linear_3d.pdf}
			\label{fig:fig_linear_3d}
		}
		\caption{Different views of Linear Projection of Flattened Patches in Video Encoder. Dotted boxes with color are kernels.}\label{fig_linear}
	\end{figure}

	The ViT \cite{dosovitskiy2021an} first extracts non-overlapping image patches, then performs a linear projection to project them into 1D tokens, and exploits the transformer architecture to model the interaction between each patch of the input image to get the final representation. Following the ViT and CLIP, we use the output from the [\texttt{class}] token as the image representation. For the input frame sequence of video $v_i=\{v^1_i, v^2_i, \dots, v^{|v_i|}_i\}$, the generated representation can denote as $\mathbf{Z}_i=\{\mathbf{z}^1_i, \mathbf{z}^2_i, \dots, \mathbf{z}^{|v_i|}_i\}$.

	We investigate two types of linear projections in the Linear Projection of Flattened Patches module shown in Figure \ref{fig:structure} named 2D linear and 3D linear separately. a) The linear projection of flattened patches of ViT is regarded as 2D linear, which embeds each 2D frame patch independently. Such a 2D linear ignores the temporal information among frames. b) Therefore, we investigate a 3D linear projection, similar to \cite{arnab2021vivit}, to enhance temporal feature extraction. The comparison between 2D and 3D is shown in Figure \ref{fig_linear}. The 3D linear embeds patches across time. Concretely, the 3D linear uses a 3D convolution with kernel $[t \times h \times w]$ as the linear instead of the kernel of $[h \times w]$ in 2D linear, where $t, h$, and $w$ are temporal, height, and width dimensions, respectively.

	\subsection{Text Encoder} We directly apply the text encoder from the CLIP to generate the caption representation. The text encoder is a Transformer \cite{vaswani2017attention} with the architecture modifications described in \cite{radford2019language}. It is a 12-layer 512-wide model with 8 attention heads. Following CLIP, the activations from the highest layer of the transformer at the [\texttt{EOS}] token are treated as the feature representation of the caption. For the caption $t_j \in \mathcal{T}$, we denote the representation as $\mathbf{w}_j$.

	\subsection{Similarity Calculator}
	\label{similarity_calculator}
	After extracting the video representation $\mathbf{Z}_i=\{\mathbf{z}^1_i, \mathbf{z}^2_i, \dots, \mathbf{z}^{|v_i|}_i\}$ and caption representation $\mathbf{w}_j$, the key point comes to the similarity calculation. Since our model is built based upon a pre-trained image-text model, we should carefully add new learnable weights in the similarity calculator module. It is hard to learn without weight initialization and may hurt the performance of the pre-trained model training with backpropagation. Therefore, we categorize the mechanisms of the similarity calculator into three categories depending on whether the module introduces new parameters to learn. The \emph{parameter-free} approach, i.e., meaning pooling, fuses the video representation without new parameters. Additionally, two other approaches introduce new weights to learn including a \emph{sequential} type and a \emph{tight} type methods with different sizes of new weights. Figure \ref{fig:similarity_calculator} illustrates the detailed structure of the three mechanisms. The parameter-free type and sequential type similarity calculators belong to the loose type that adopts two separate branches for video and text representation independently to calculate cosine similarity. While the tight type similarity calculator uses the transformer model for multi-modal interaction and further calculates the similarity via a linear projection, both of which consist of new weights to learn.

	\paragraph{Parameter-free type}
	According to the CLIP, the frames representation $\mathbf{Z}_i$ and the caption representation $\mathbf{w}_j$ have been layer normalized and linearly projected into a multi-modal embedding space through the large-scale pretraining on the image-text pairs. The natural idea is to employ a parameter-free type to calculate similarity directly with the image/frame from the video perspective. The parameter-free type first uses the mean pooling to aggregate the feature of all frames to obtain an `average frame', $\hat{\mathbf{z}}_i = \texttt{mean-pooling}(\mathbf{z}^1_i, \mathbf{z}^2_i, \dots, \mathbf{z}^{|v_i|}_i)$. Then, the similarity function $s(v_i, t_j)$ is defined as the cosine similarity,
	\begin{align}
		s(v_i, t_j) = \frac{{\mathbf{w}_j}^\top \hat{\mathbf{z}}_i}{\lVert \mathbf{w}_j \rVert \lVert \hat{\mathbf{z}}_i \rVert}. \label{eq:cosine_sim}
	\end{align}

	\paragraph{Sequential type}
	The mean-pooling operation ignores the sequential information between frames. In this way, we explore two methods to model the sequential feature for Sequential type similarity calculator. One is LSTM \cite{hochreiter1997long, gers2002learning}, and the other one is Transformer encoder \cite{vaswani2017attention} with position embedding $\mathbf{P}$, Both of which are effective models for sequence features. We formulate them as $\tilde{\mathbf{Z}}_i = \texttt{LSTM}(\mathbf{Z}_i)$ and $\tilde{\mathbf{Z}}_i = \texttt{Transformer-Enc}(\mathbf{Z}_i + \mathbf{P})$, respectively. Through the encoding, the $\tilde{\mathbf{Z}}_i$ already embeds the temporal information. The subsequent operations are the same as the parameter-free type similarity calculator, and the similarity function is also the Eq. (\ref{eq:cosine_sim}), and $\hat{\mathbf{z}}_i = \texttt{mean-pooling}(\tilde{\mathbf{Z}}_i)$.

	\paragraph{Tight type}
	Different from above parameter-free type and sequential type, the tight type uses a Transformer Encoder \cite{vaswani2017attention} for multimodal interaction between video and caption similar to \cite{Luo2020UniVL}, and predict the similarity through a linear layer, which introduces the most uninitialized weights. First, the Transformer Encoder takes the concatenated caption representation $\mathbf{w}_j$ and frames' representation $\mathbf{Z}_i=\{\mathbf{z}^1_i, \mathbf{z}^2_i, \dots, \mathbf{z}^{|v_i|}_i\}$ as the fused feature $\mathbf{U}_i$ formulated as,
	\begin{align}
		\mathbf{U}_i &= [\mathbf{w}_j, \mathbf{z}^1_i, \mathbf{z}^2_i, \dots, \mathbf{z}^{|v_i|}_i],\\
		\tilde{\mathbf{U}}_i &= \texttt{Transformer-Enc}(\mathbf{U}_i + \mathbf{P} + \mathbf{T}).
	\end{align}
	where $[,]$ denotes concatenate operation. $\mathbf{P}$ is the position embedding, and $\mathbf{T}$ is type embedding similar to Segment
	Embeddings in BERT \cite{devlin2019bert}. The $\mathbf{T}$ contains two types of embedding, one is for caption and the other is for video frames. Next, we calculate similarity score with two linear projection layers plus an activation upon the first token output of the last layer $\tilde{\mathbf{U}}_i[0,:]$. Concretely, the similarity function $s(v_i, t_j) = \texttt{FC}\big(\texttt{ReLU}\big(\texttt{FC}(\tilde{\mathbf{U}}_i[0,:])\big)\big)$, where \texttt{FC} is the linear projection, and the \texttt{ReLU} means ReLU activation function \cite{Agarap2018ReLU}.

	\subsection{Training Strategy}
	\paragraph{Loss Function}
	\begin{table*}[!t]
		\begin{subtable}{.5\linewidth}
			\vspace{.91cm}
			\setlength{\tabcolsep}{2pt}
			\centering
			\scalebox{0.70}{
				\begin{tabular}{lccccccc}
					\toprule 
					Methods        & TrainD & E2E & R@1$\uparrow$   & R@5$\uparrow$ & R@10$\uparrow$ & MdR$\downarrow$ & MnR$\downarrow$  \\ \midrule
					C+LSTM+SA$^a$			& M & \checkmark & 4.2 & 12.9 & 19.9 & 55 & - \\  
					VSE$^b$					& M & \checkmark & 3.8 & 12.7 & 17.1 & 66 & - \\  
					SNUVL$^c$				& M & \checkmark & 3.5 & 15.9 & 23.8 & 44 & - \\  
					Kaufman et al.$^d$		& M & \checkmark & 4.7 & 16.6 & 24.1 & 41 & - \\ 
					CT-SAN$^e$				& M & \checkmark & 4.4 & 16.6 & 22.3 & 35 & - \\ 
					JSFusion$^f$			& M & \checkmark & 10.2 & 31.2 & 43.2 & 13 & - \\ 
					HowTo100M$^g$			& H+M & \checkmark & 14.9 & 40.2 & 52.8 & 9 & - \\ 
					ActBERT$^h$				& H+M &  & 8.6 & 23.4 & 33.1 & 36 & - \\ 
					NoiseE$^i$				& H+M &  & 17.4 & 41.6 & 53.6 & 8 & - \\
					UniVL$^j$				& H+M &  & 21.2 & 49.6 & 63.1 & 6 & - \\
					HERO$^k$				& H+M &  & 16.8 & 43.4 & 57.7 & - & - \\
					ClipBERT$^l$			& C+G+M & \checkmark & 22.0 & 46.8 & 59.9 & 6 & - \\
					\midrule
					(Ours)-meanP					& W+M & \checkmark & \textbf{42.1} & \textbf{71.9} & \textbf{81.4} & \textbf{2} & \textbf{15.7} \\
					(Ours)-seqLSTM					& W+M & \checkmark & 41.7 & 68.8 & 78.7 & \textbf{2} & 16.6 \\
					(Ours)-seqTransf				& W+M & \checkmark & 42.0 & 68.6 & 78.7 & \textbf{2} & 16.2 \\
					(Ours)-tightTransf				& W+M & \checkmark & 37.8 & 68.4 & 78.4 & \textbf{2} & 17.2 \\
					\bottomrule
				\end{tabular}
			}
			\caption{Training on Training-7K}
		\end{subtable}%
		\begin{subtable}{.5\linewidth}
			\begin{subtable}{1.\linewidth}
				\setlength{\tabcolsep}{2pt}
				\centering
				\scalebox{0.70}{
					\begin{tabular}{lccccccc}
						\toprule 
						Methods        & TrainD & E2E & R@1$\uparrow$   & R@5$\uparrow$ & R@10$\uparrow$ & MdR$\downarrow$ & MnR$\downarrow$  \\ \midrule
						MIL-NCE$^m$				& H & \checkmark & 9.9 & 24.0 & 32.4 & 29.5 & - \\ 
						CLIP-straight$^n$				& W & \checkmark & 31.2 & 53.7 & 64.2 & 4 & - \\
						\bottomrule
					\end{tabular}
				}
				\caption{Zero-shot}
			\end{subtable} 
			\newline
			\vspace{.3cm}
			\newline
			\begin{subtable}{1.\linewidth}
				\setlength{\tabcolsep}{2pt}
				\centering
				\scalebox{0.70}{
					\begin{tabular}{lccccccc}
						\toprule 
						Methods        & TrainD & E2E & R@1$\uparrow$   & R@5$\uparrow$ & R@10$\uparrow$ & MdR$\downarrow$ & MnR$\downarrow$  \\ \midrule
						CE$^o$					& M & & 20.9 & 48.8 & 62.4 & 6 & 28.2 \\ 
						MMT$^p$					& H+M & & 26.6 & 57.1 & 69.6 & 4 & 24.0 \\
						AVLnet$^q$				& H+M & & 27.1 & 55.6 & 66.6 & 4 & - \\
						SSB$^r$					& H+M & & 30.1 & 58.5 & 69.3 & 3 & - \\
						MDMMT$^s$				& MD+M & & 38.9 & 69.0 & 79.7 & \textbf{2} & 16.5 \\
						Frozen$^t$				& CW+M & \checkmark & 31.0 & 59.5 & 70.5 & 3 & - \\
						HiT$^u$					& H+M &  & 30.7 & 60.9 & 73.2 & 2.6 & - \\
						TT-CE+$^v$					& M &  & 29.6 & 61.6 & 74.2 & 3 & - \\
						\midrule
						(Ours)-meanP					& W+M & \checkmark & 43.1 & 70.4 & 80.8 & \textbf{2} & 16.2 \\
						(Ours)-seqLSTM					& W+M & \checkmark & 42.5 & 70.8 & 80.7 & \textbf{2} & 16.7 \\
						(Ours)-seqTransf				& W+M & \checkmark & \textbf{44.5} & 71.4 & \textbf{81.6} & \textbf{2} & 15.3 \\
						(Ours)-tightTransf				& W+M & \checkmark & 40.2 & \textbf{71.5} & 80.5 & \textbf{2} & \textbf{13.4} \\
						\bottomrule
					\end{tabular}
				}
				\caption{Training on Training-9K}
			\end{subtable} 
		\end{subtable} 
		\caption{Results of text-to-video retrieval on MSR-VTT dataset. Table (a) and (c) present the results on different splits of the dataset. `Training-7K' follows the data splits from \cite{miech2019howto100m} and `Training-9K' follows the data splits from \cite{Gabeur2020MMT}. They have the same test set but different training sets. For each table, the column `TrainD' shows the datasets used for pre-training and training, where M, H, W, C, G denote MSR-VTT, HowTo100M \cite{miech2019howto100m}, WIT \cite{radford2021learning}, COCO Captions \cite{Chen2015COCO}, and Visual Genome Captions \cite{Krishna2017Genome}. Besides, MD used in MDMMT\cite{Dzabraev2021MDMMT} denotes a combined multidomain dataset including MSR-VTT, LSMDC, HowTo100M, etc., and CW means CC3M \cite{sharma2018conceptual} plus WebVid-2M \cite{Bain2021Frozen}. The column `E2E' with \checkmark means training from raw video in an end-to-end manner. The baseline methods are $^a$C+LSTM+SA \cite{Torabi2016Learning}, $^b$VSE \cite{kiros2014unifying}, $^c$SNUVL \cite{Yu2016VideoCaptioning}, $^d$Kaufman et al. \citet{Kaufman2017Temporal}, $^e$CT-SAN \cite{Yu2017End}, $^f$JSFusion \cite{yu2018joint}, $^g$HowTo100M \cite{miech2019howto100m}, $^h$ActBERT \cite{Zhu_2020_CVPR}, $^i$NoiseE \cite{Amrani2020Noise}, $^j$UniVL \cite{Luo2020UniVL}, $^k$HERO \cite{Li2020HERO}, $^l$ClipBERT \cite{lei2021less}, $^m$MIL-NCE \cite{miech19endtoend}, $^n$CLIP-straight \cite{PortilloQuintero2021}, $^o$CE \cite{Liu2019CE}, $^p$MMT \cite{Gabeur2020MMT}, $^q$AVLnet \cite{Rouditchenko2020}, $^r$SSB \cite{patrick2021supportset}, $^s$MDMMT \cite{Dzabraev2021MDMMT}, $^t$Frozen \cite{Bain2021Frozen}, $^u$HiT \cite{Liu2021HiT}, $^v$TT-CE+ \cite{croitoru2021teachtext}.}
		\label{tab:result_of_retrieval_MSR-VTT}
	\end{table*}
	Given a batch of $B$ (video, text) or (video clip, text) pairs, the model needs to generate and optimize $B \times B$ similarities. We use a symmetric cross entropy loss over these similarity scores to train the model's parameters,
	\begin{align}
		\mathcal{L}_{v2t} &= -\frac{1}{B} \sum_i^B{\log \frac{\exp(s(v_i, t_i))}{\sum_{j=1}^B{\exp(s(v_i, t_j)}}}, \\
		\mathcal{L}_{t2v} &= -\frac{1}{B} \sum_i^B{\log \frac{\exp(s(v_i, t_i))}{\sum_{j=1}^B{\exp(s(v_j, t_i)}}}, \\
		\mathcal{L} &= \mathcal{L}_{v2t} + \mathcal{L}_{t2v}.
	\end{align}
	The loss $\mathcal{L}$ is the sum of video-to-text loss $\mathcal{L}_{v2t}$ and text-to-video loss $\mathcal{L}_{t2v}$.

	\paragraph{Frame Sampling}
	Since our model is trained on pixels directly via taking the frames as input, it is an important strategy to extract frames. An effective sampling strategy is required to consider the balance between the information richness and the computational complexity (especially memory cost). To consider the sequential information in the video (or video clip), we adopt a uniform frame sampling strategy instead of a random sparse sampling strategy used in \cite{lei2021less}. The sampling rate is 1 frame per second. Besides, we also investigate different frame lengths and different extraction positions in our experiments.

	\paragraph{Pre-training}
	Although the CLIP is effective for learning the visual concepts of images, it is essential to learn temporal features from video. To further transfer the knowledge to video, our CLIP4Clip model is post-pretrained on Howto100M dataset \cite{miech2019howto100m}. Pretraining on the video-text dataset is extremely challenging due to efficiency consideration. We conduct a preliminary exploration and use the `Food and Entertaining' category, around 380k videos, as the post-pretraining dataset (called \emph{Howto100M-380k} in the rest of the paper). We adopt the MIL-NCE loss \cite{miech19endtoend} to optimize the CLIP in our parameter-free type. The optimizer is Adam \cite{kingma2014adam}, with a learning rate 1e-8. The token length is 32, the frame length is 12, and the batch size is 48. The training is processed on 8 NVIDIA Tesla V100 GPUs. We run 5 epochs and it takes about 2 weeks. In this paper, the post-pretraining test can be regarded as a preliminary study on this direction for future work.

	\section{Experiments}
	We first describe the datasets and implementation details before presenting state-of-the-art results on five datasets. We then ablate various settings of our model. Finally, we discuss some aspects of promising direction.

	\subsection{Datasets}
	We validate our model on five datasets: MSR-VTT, MSVC, LSMDC, ActivityNet, and DiDeMo.

	\noindent
	\textbf{MSR-VTT} \cite{xu2016msr} is a dataset composed of 10,000 videos, each with a length that ranges from 10 to 32 seconds and 200,000 captions. We use two types of data splits, `Training-7K' and `Training-9K', to compare with baselines. The `Training-7K' follows the data splits from \cite{miech2019howto100m} and the `Training-9K' follows the data splits from \cite{Gabeur2020MMT}. The test data in both splits is `test 1k-A', which contains 1,000 clip-text pairs following JSFusion \cite{yu2018joint}. We use `Training-9K' as the default setting if there is no extra annotation.

    \noindent
	\textbf{MSVD} \cite{chen2011collecting} contains 1,970 videos, each with a length that ranges from one to 62 seconds. Train, validation and, test splits contain 1,200, 100, and 670 videos, respectively. Each video has approximately 40 associated sentences in English.

	\noindent
	\textbf{LSMDC} \cite{Rohrbach2015LSMDC} is comprised 118,081 videos, each with a length that ranges from two to 30 seconds. The videos were extracted from 202 movies. The validation set contains 7,408 videos, and the test set 1,000 videos from movies independent from the training and validation splits.

	\noindent
	\textbf{ActivityNet} \cite{krishna2017dense} consists of 20,000 YouTube videos. We follow the setting from \cite{bowen2018cross,Gabeur2020MMT} to concatenate all the descriptions of a video to form a paragraph and evaluate the model with video-paragraph retrieval on the `val1' split.

	\noindent
	\textbf{DiDeMo} \cite{hendricks17iccv} contains 10,000 videos annotated with 40,000 sentences. We evaluate video-paragraph retrieval following \cite{Liu2019CE,lei2021less,Bain2021Frozen}, where all sentence descriptions for a video are concatenated into a single query.
	\begin{table}[!t]
		\setlength{\tabcolsep}{2pt}
		\centering
		\scalebox{0.70}{
			\begin{tabular}{lccccccc}
				\toprule
				Methods  & TrainD & E2E & R@1$\uparrow$   & R@5$\uparrow$ & R@10$\uparrow$ & MdR$\downarrow$ & MnR$\downarrow$  \\ \midrule
				Multi Cues$^a$ & M & \checkmark & 20.3 & 47.8 & 61.1 & 6  & -\\ 
				CE$^b$   & M & & 19.8 & 49.0 & 63.8 & 6  & -\\ 
				SSB$^c$ & H+M & & 28.4 & 60.0 & 72.9 & 4  & -\\  
				NoiseE$^d$		& H+M  & & 20.3 & 49.0 & 63.3 & 6 & - \\
				CLIP-straight$^e$   & W & \checkmark & 37.0 & 64.1 & 73.8 & 3  & -\\ 
				Frozen$^f$	& CW+M & \checkmark & 33.7 & 64.7 & 76.3 & 3 & - \\
				TT-CE+$^g$	& M & & 25.4 & 56.9 & 71.3 & 4 & - \\
				\midrule
				(Ours)-meanP					& W+M & \checkmark & \textbf{46.2} & \textbf{76.1} & \textbf{84.6} & \textbf{2} & \textbf{10.0} \\
				(Ours)-seqLSTM					& W+M & \checkmark & \textbf{46.2} & 75.3 & 84.5 & \textbf{2} & 10.2 \\
				(Ours)-seqTransf				& W+M & \checkmark & 45.2 & 75.5 & 84.3 & \textbf{2} & 10.3 \\
				(Ours)-tightTransf				& W+M & \checkmark & 40.0 & 71.5 & 82.1 & \textbf{2} & 13.3 \\
				\bottomrule
			\end{tabular}
		}
		\caption{Results of text-to-video retrieval on MSVD dataset. In the column `TrainD', M, H, and W denote training on MSVD, HowTo100M \cite{miech2019howto100m}, and WIT \cite{radford2021learning}, and CW means CC3M \cite{sharma2018conceptual} plus WebVid-2M \cite{Bain2021Frozen}. The column `E2E' with \checkmark means training from raw video in an end-to-end manner. The baseline methods are $^a$Multi Cues \cite{Mithun2018Learning}, $^b$CE \cite{Liu2019CE}, $^c$SSB \cite{patrick2021supportset}, $^d$NoiseE \cite{Amrani2020Noise}, $^e$CLIP-straight \cite{PortilloQuintero2021}, $^f$Frozen \cite{Bain2021Frozen}, $^g$TT-CE+ \cite{croitoru2021teachtext}.}
		\label{tab:result_of_retrieval_msvd}
	\end{table}
	\begin{table}[!t]
		\setlength{\tabcolsep}{2pt}
		\centering
		\scalebox{0.70}{
			\begin{tabular}{lccccccc}
				\toprule
				Methods  & TrainD & E2E & R@1$\uparrow$   & R@5$\uparrow$ & R@10$\uparrow$ & MdR$\downarrow$ & MnR$\downarrow$  \\ \midrule
				CT-SAN$^a$ 		& L & \checkmark & 5.1 & 16.3 & 25.2 & 46.0 & - \\ 
				JSFusion$^b$  & L & \checkmark & 9.1 & 21.2 & 34.1 & 36.0  & -\\  
				CE$^c$   & L & & 11.2 & 26.9 & 34.8 & 25.3  & 96.8\\ 
				MMT$^d$   & H+L & & 12.9 & 29.9 & 40.1 & 19.3  & 75.0\\ 
				NoiseE$^e$		& H+L &  & 6.4 & 19.8 & 28.4 & 39.0 & - \\
				CLIP-straight$^f$   & L & \checkmark & 11.3 & 22.7 & 29.2 & 56.5  & -\\ 
				MDMMT$^g$  & MD+L & & 18.8 & 38.5 & 47.9 & 12.3 & \textbf{58.0} \\ 
				Frozen$^h$	& CW+L & \checkmark & 15.0 & 30.8 & 39.8 & 20.0 & - \\
				HiT$^i$		& H+L &  & 14.0 & 31.2 & 41.6 & 18.5 & - \\
				TT-CE+$^j$	& L & & 17.2 & 36.5 & 46.3 & 13.7 & - \\
				\midrule
				(Ours)-meanP					& W+L & \checkmark & 20.7 & 38.9 & 47.2 & 13.0 & 65.3 \\
				(Ours)-seqLSTM					& W+L & \checkmark & 21.6 & \textbf{41.8} & \textbf{49.8} & \textbf{11.0} & \textbf{58.0} \\
				(Ours)-seqTransf				& W+L & \checkmark & \textbf{22.6} & 41.0 & 49.1 & \textbf{11.0} & 61.0 \\
				(Ours)-tightTransf				& W+L & \checkmark & 18.9 & 37.8 & 46.7 & 13.0 & 61.6 \\
				\bottomrule
			\end{tabular}
		}
		\caption{Results of text-to-video retrieval on LSMDC dataset. In the column `TrainD', L, H, and W denote training on LSMDC, HowTo100M \cite{miech2019howto100m}, and WIT \cite{radford2021learning}, MD used in \cite{Dzabraev2021MDMMT} denotes a combined multidomain dataset containing MSR-VTT, LSMDC, HowTo100M, etc., and CW means CC3M \cite{sharma2018conceptual} plus WebVid-2M \cite{Bain2021Frozen}. The column `E2E' with \checkmark means training from raw video in an end-to-end manner. The baseline methods are $^a$CT-SAN \cite{Yu2017End}, $^b$JSFusion \cite{yu2018joint}, $^c$CE \cite{Liu2019CE}, $^d$MMT \cite{Gabeur2020MMT}, $^e$NoiseE \cite{Amrani2020Noise}, $^f$CLIP-straight \cite{PortilloQuintero2021}, $^g$MDMMT \cite{Dzabraev2021MDMMT}, $^h$Frozen \cite{Bain2021Frozen}, $^i$HiT \cite{Liu2021HiT}, $^j$TT-CE+ \cite{croitoru2021teachtext}.}
		\label{tab:result_of_retrieval_lsmdc}
	\end{table}
	\begin{table}[!t]
		\setlength{\tabcolsep}{2pt}
		\centering
		\scalebox{0.70}{
			\begin{tabular}{lccccccc}
				\toprule
				Methods  & TrainD & E2E & R@1$\uparrow$   & R@5$\uparrow$ & R@50$\uparrow$ & MdR$\downarrow$ & MnR$\downarrow$  \\ \midrule
				FSE$^a$	 & A & & 18.2 & 44.8 & 89.1 & 7.0 & - \\ 
				CE$^b$   & A & & 18.2 & 47.7 & 91.4 & 6.0 & 23.1\\ 
				HSE$^a$  & A & & 20.5 & 49.3 & - & -  & -\\ 
				MMT$^c$   & H+A & & 28.7 & 61.4 & 94.5 & 3.3 & 16.0\\ 
				SSB$^d$ & H+A & & 29.2 & 61.6 & 94.7 & 3.0 & -\\  
				HiT$^e$		& H+A &  & 29.6 & 60.7 & 95.6 & 3.0 & - \\
				ClipBERT$^f$			& C+G+A & \checkmark & 21.3 & 49.0 & - & 6.0 & - \\
				TT-CE+$^g$	& A & & 23.5 & 57.2 & 96.1 & 4.0 & - \\
				\midrule
				(Ours)-meanP					& W+A & \checkmark & \textbf{40.5} & \textbf{72.4} & 98.1 & \textbf{2.0} & 7.4 \\
				(Ours)-seqLSTM					& W+A & \checkmark & 40.1 & 72.2 & 98.1 & \textbf{2.0} & \textbf{7.3} \\
				(Ours)-seqTransf				& W+A & \checkmark & \textbf{40.5} & \textbf{72.4} & \textbf{98.2} & \textbf{2.0} & 7.5 \\
				(Ours)-tightTransf				& W+A & \checkmark & 19.5 & 47.6 & 93.1 & 6.0 & 17.3 \\
				\bottomrule
			\end{tabular}
		}
		\caption{Results of text-to-video retrieval on ActivityNet dataset. In the column `TrainD', A, H, W, C, and G denote training on ActivityNet, HowTo100M \cite{miech2019howto100m}, WIT \cite{radford2021learning}, COCO Captions \cite{Chen2015COCO}, and Visual Genome Captions \cite{Krishna2017Genome}. The column `E2E' with \checkmark means training from raw video in an end-to-end manner. The baseline methods are $^a$FSE,HSE \cite{bowen2018cross}, $^b$CE \cite{Liu2019CE}, $^c$MMT \cite{Gabeur2020MMT}, $^d$SSB \cite{patrick2021supportset}, $^e$HiT \cite{Liu2021HiT}, $^f$ClipBERT \cite{lei2021less}, $^g$TT-CE+ \cite{croitoru2021teachtext}.}
		\label{tab:result_of_retrieval_activityNet}
	\end{table}
	\begin{table}[!t]
		\setlength{\tabcolsep}{2pt}
		\centering
		\scalebox{0.70}{
			\begin{tabular}{lccccccc}
				\toprule
				Methods  & TrainD & E2E & R@1$\uparrow$   & R@5$\uparrow$ & R@10$\uparrow$ & MdR$\downarrow$ & MnR$\downarrow$  \\ \midrule
				S2VT$^a$      & D & \checkmark & 11.9 & 33.6 & - & 13.0 & - \\ 
				FSE$^b$	      & D & & 13.9 & 36.0 & - & 11.0 & - \\ 
				CE$^c$        & D & & 16.1 & 41.1 & - & 8.3  & 43.7\\ 
				ClipBERT$^d$$^\dagger$  & C+G+D & \checkmark & 20.4 & 48.0 & 60.8 & 6.0 & - \\
				Frozen$^e$$^\dagger$	  & CW+D & \checkmark & 34.6 & 65.0 & 74.7 & 3.0 & - \\
				TT-CE+$^f$	& D & & 21.6 & 48.6 & 62.9 & 6.0 & - \\
				\midrule
				(Ours)-meanP					& W+D & \checkmark & \textbf{43.4} & \textbf{70.2} & \textbf{80.6} & \textbf{2.0} & \textbf{17.5} \\
				(Ours)-seqLSTM					& W+D & \checkmark & \textbf{43.4} & 69.9 & 80.2 & \textbf{2.0} & \textbf{17.5} \\
				(Ours)-seqTransf				& W+D & \checkmark & 42.8 & 68.5 & 79.2 & \textbf{2.0} & 18.9 \\
				(Ours)-tightTransf				& W+D & \checkmark & 25.8 & 52.8 & 66.3 & 5.0 & 27.3 \\
				\bottomrule
			\end{tabular}
		}
		\caption{Results of text-to-video retrieval on DiDeMo dataset. In the column `TrainD', D, H, W, C, and G denote training on DiDeMo, HowTo100M \cite{miech2019howto100m}, WIT \cite{radford2021learning}, COCO Captions \cite{Chen2015COCO}, and Visual Genome Captions \cite{Krishna2017Genome}, CW means CC3M \cite{sharma2018conceptual} plus WebVid-2M \cite{Bain2021Frozen}. The column `E2E' with \checkmark means training from raw video in an end-to-end manner. $\dagger$ means that the candidate video is concatenated using ground truth proposals. The baseline methods are $^a$S2VT \cite{Venugopalan2015Translating}, $^b$FSE \cite{bowen2018cross}, $^c$CE \cite{Liu2019CE}, $^d$ClipBERT \cite{lei2021less}, $^e$Frozen \cite{Bain2021Frozen}, $^f$TT-CE+ \cite{croitoru2021teachtext}.}
		\label{tab:result_of_retrieval_didemo}
	\end{table}

	We use standard retrieval metrics: recall at rank $K$ (R@$K$, higher is better), median rank (MdR, lower is better), and mean rank (MnR, lower is better) to evaluate the performance of our model. R@$K$ (Recall at $K$) calculates the percentage of test samples for which the correct result is found in the top-$K$ retrieved points to the query sample. We report results for R@1, R@5, and R@10 (or R@50 for the ActivityNet). Median Rank calculates the median of the ground-truth results in the ranking. Similarly, Mean Rank calculates the mean rank of all correct results.
	\begin{figure*}[tbp]
		\centering
		\subfloat[Batch size] {
			\centering
			\includegraphics[width=0.23\textwidth]{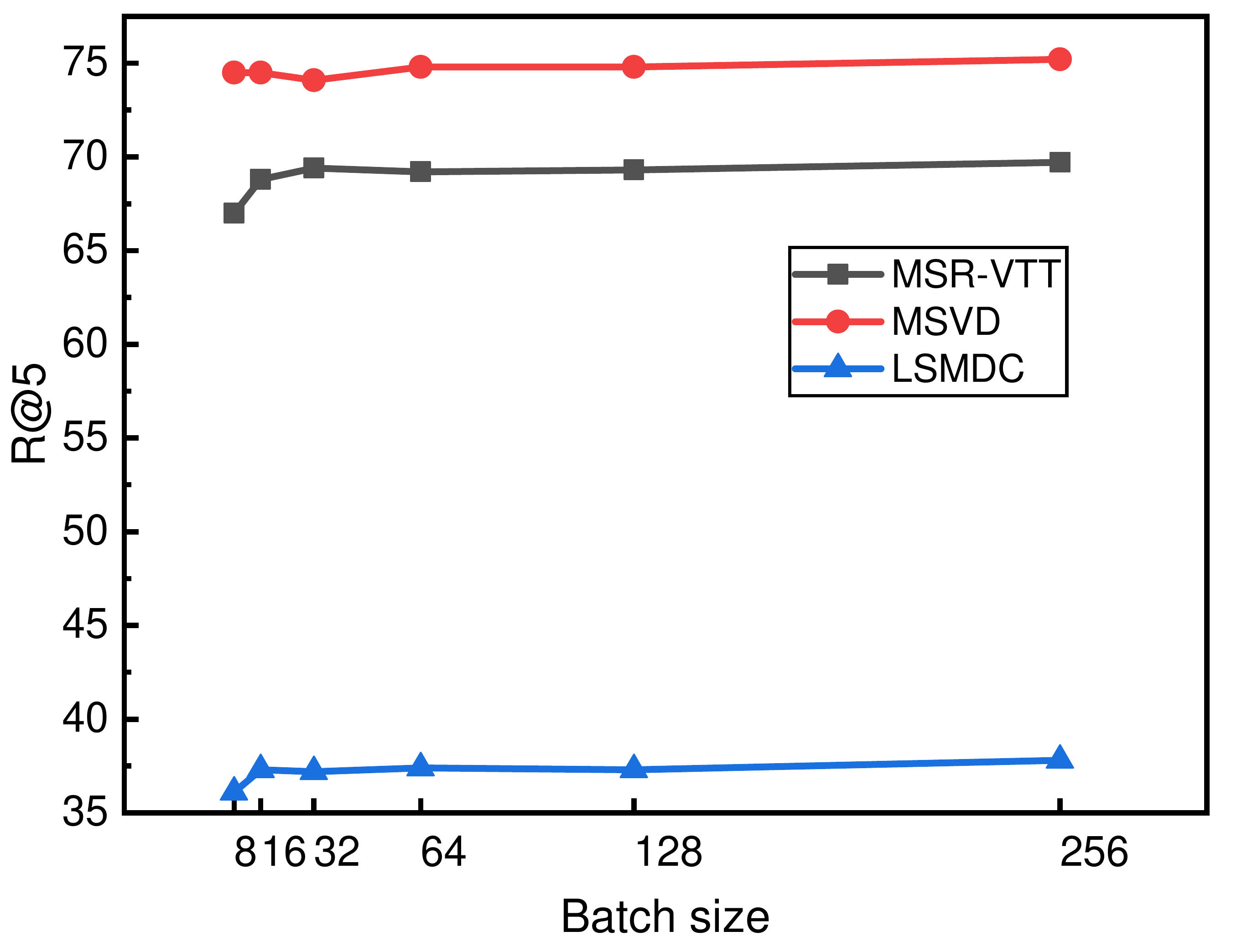}
			\label{fig:batch_size}
		}
		\subfloat[Learning rate] {
			\centering
			\includegraphics[width=0.23\textwidth]{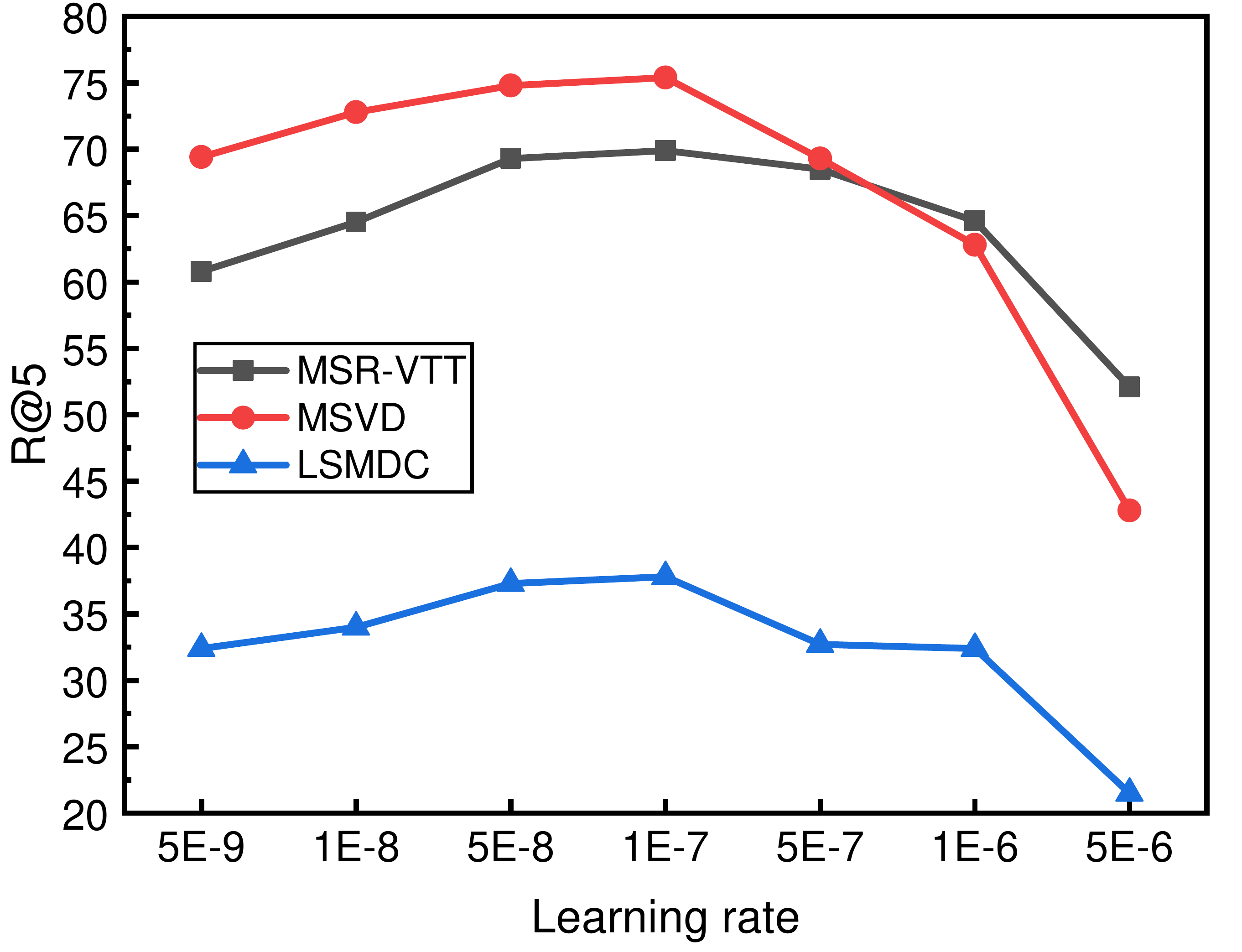}
			\label{fig:lr}
		}
		\subfloat[Freeze layer] {
			\centering
			\includegraphics[width=0.23\textwidth]{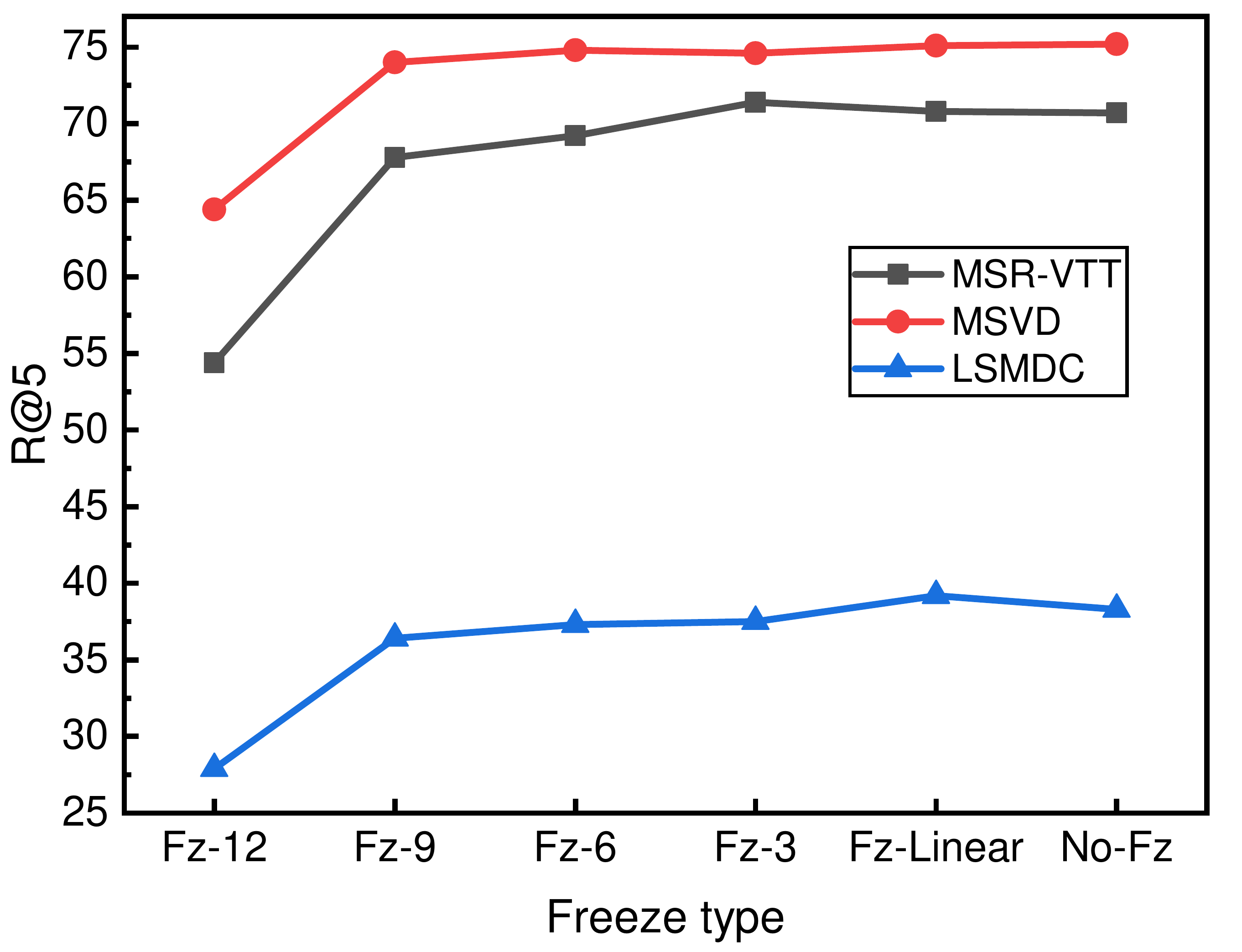}
			\label{fig:freeze_layer}
		}
		\subfloat[Frame length] {
			\centering
			\includegraphics[width=0.23\textwidth]{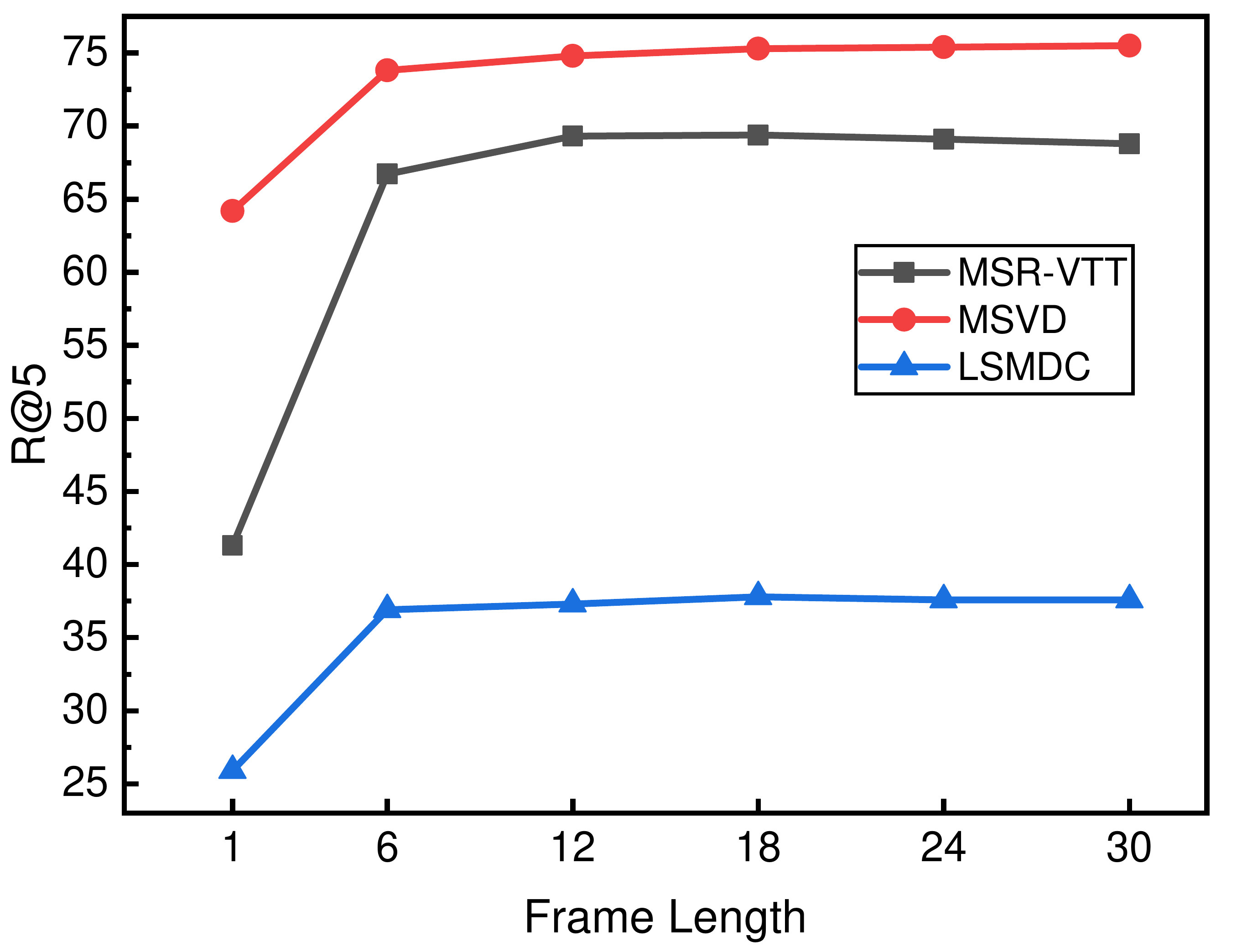}
			\label{fig:frame_length}
		}
		\caption{Retrieval results on different batch sizes, frame length, freeze layer, and learning rate. \textbf{Batch size}: freeze layer is 6. \textbf{Learning rate}: batch size is 128, freeze layer is 6, frame length is 12. \textbf{Freeze layer}: Fz-[NO.] means freeze layers below [NO.]-th layer (inclusive), Fz-Linear means only freeze the linear layer at the bottom, No-Fz is training without freeze, batch size is 128, frame length is 12, learning rate is 5e-8. \textbf{Frame length}: batch size is 128, freeze layer is 6, learning rate is 5e-8.}\label{fig_study_sdy_fig}
	\end{figure*}

	\subsection{Experimental Details}
	We initial the text encoder and video encoder with CLIP (ViT-B/32) \cite{radford2021learning} in this paper. The practical
	question is how to initialize the parameters in the similarity calculator, e.g., parameters in sequential type. Our solution is to reuse similar parameters from the CLIP (ViT-B/32). Concretely, for the position embedding in sequential type and tight type, we initialize them by repeating the position embedding from CLIP's text encoder. Similarly, the transformer encoder is initialized by the corresponding layers' weight of the pretrained CLIP's image encoder. The rest of the parameters, e.g., LSTM and linear projection, are initialized randomly. The temporal dimension $t$, height dimension $h$, and width dimension $w$ of 3D linear and 2D linear in Section \ref{sec:video_encoder} are set to 3, 32, 32, respectively. For 3D linear, we set stride and padding with 1 at the temporal dimension. We initialize the 3D linear following \cite{arnab2021vivit} with a `central frame initialization' strategy from the pretrained 2D linear of CLIP. Concretely, we use $[\mathbf{0}, \mathbf{E}_{2D}, \mathbf{0}]$ from the 2D weight $\mathbf{E}_{2D}$ of CLIP' image encoder.

	We finetune the model with the Adam optimizer \cite{kingma2014adam}. For the learning rate, we decay it using a cosine schedule \cite{Loshchilov2017SGDR} following the CLIP \cite{radford2021learning}. If no otherwise specified, the initial learning rate is 1e-7 for text encoder and video encoder (including linear projection) and 1e-4 for new modules, e.g., LSTM, the caption token length is 32, the frame length is 12, the batch size is 128, and running 5 epochs. The layer of LSTM is 1, and the layer of Transformer Encoder in both sequential type and tight type is 4 in our experiments. All finetune experiments are carried out on 4 NVIDIA Tesla V100 GPUs. Note that the ActivityNet and the DiDeMo are regarded as video-paragraph retrieval, so we set the caption token length and the frame length 64. The experiments on them are carried out on 16 NVIDIA Tesla V100 GPUs.

	\subsection{Comparison to the State of the Art}
	We compare all types of similarity calculator based on the pretrained CLIP against the state-of-the-art (SOTA): `-meanP', `-seqLSTM', `-seqTransf', and `-tightTransf' are short for parameter-free type (i.e., mean pooling), sequential type of LSTM, Transformer Encoder, and tight type mentioned in Section \ref{similarity_calculator}. Table \ref{tab:result_of_retrieval_MSR-VTT}-\ref{tab:result_of_retrieval_didemo} present the text-to-video retrieval results of our model on MSR-VTT, MSVC, LSMDC, ActivityNet, and DiDeMo. The baselines of each dataset are listed in the caption of each table for clarification. We achieve the SOTA results on all five datasets by a large margin compared with all baselines. We find that the growth of retrieval performance benefits from the pretrained CLIP via our results and the concurrent CLIP-straight \cite{PortilloQuintero2021}. Besides, the improvement from our end-to-end finetune proves the potential of the image-text pretrain model on video-text retrieval.

	For the MSR-VTT dataset, the model with the parameter-free type (-meanP) achieves the best results for the `Training-7k' data split, while the model with the sequential type (-seqTransf) outperforms other methods for the `Training-9k' data split. We think that it is hard to learn extra parameters beyond the pre-trained parameters given a small dataset. With a large dataset, it is capable of learning the extra parameters. For the LSMDC dataset, the model with the sequential type performs better than the other two types. The two sequential types, -seqLSTM and -seqTransf, achieve comparable results. For the MSVD dataset, the performance of the parameter-free type is the best. We notice that the MSVD training data is smaller than the MSR-VTT and MSVD datasets at least 2 times, and consider the reason is that the extra parameters need extra large dataset to keep the advance from pretrained weight. The performance of video-paragraph retrieval on the ActivityNet and DiDeMo further proves the advantage of the parameter-free type when utilizing the pretrained model. Among five datasets, almost all the results of the tight type (-tightTransf) are the worst among all calculators. We think that the tight type is still hard to learn the cross-modality interaction without enough dataset.
	\begin{table}[!t]
		\setlength{\tabcolsep}{5pt}
		\centering
		\scalebox{0.75}{
		\begin{tabular}{cccccc}
			\toprule
			Frame Selection & R@1$\uparrow$   & R@5$\uparrow$ & R@10$\uparrow$ & MdR$\downarrow$ & MnR$\downarrow$  \\ \midrule
			\multicolumn{6}{c}{MSR-VTT} \\
			Head & 42.3 & \textbf{70.8} & \textbf{80.8} & 2 & \textbf{15.7} \\
			Tail & 40.5 & 67.9 & 77.4 & 2 & 18.8 \\
			Uniform & \textbf{42.6} & 70.4 & 80.1 & 2 & 16.3 \\
			\midrule
			\multicolumn{6}{c}{MSVD} \\
			Head & 45.7 & 75.1 & 84.1 & 2 & 10.2 \\
			Tail & 45.6 & \textbf{75.3} & 84.3 & 2 & 10.2 \\
			Uniform & \textbf{46.0} & \textbf{75.3} & \textbf{84.5} & 2 & \textbf{10.1} \\
			\midrule
			\multicolumn{6}{c}{LSMDC} \\
			Head & 20.3 & 39.2 & 46.8 & 13 & \textbf{63.3} \\
			Tail & \textbf{20.7} & 38.2 & 46.4 & 14 & 63.6 \\
			Uniform & \textbf{20.7} & \textbf{39.0} & \textbf{47.1} & 13 & 63.4 \\
			\bottomrule
		\end{tabular}
		}
		\caption{Study on sampling strategy. `Head', `Tail' and `Uniform' are three sample strategies to select frames from a video. Batch size is 128, freeze layer is 0, frame length is 12, and learning rate is 5e-8.}
		\label{tab:study_of_retrieval_onorder_MSR-VTT}
	\end{table}

	\subsection{Hyperparameters and Learning Strategy}
	We run extensive experiments on studying the hyperparameters and learning strategies to search for the best settings. Figure \ref{fig_study_sdy_fig} shows the resulting chart. With increasing of \emph{Batch size} in \ref{fig:batch_size}, the performance increases and it achieves comparable result for batch size 128 and 256. In our experiment, we set the batch size to 128. About the study on \emph{frame length} in \ref{fig:frame_length}, we can see a significant increase between 1 and 6 frames which shows that it is required for video to actually model with a sequence of multiple frames instead of one single frame. We sampled 12 frames for our experiment, which is both efficient and effective. We also study whether we should \emph{freeze} the parameters of each layer pre-trained by CLIP. From the Figure \ref{fig:freeze_layer}, it is better to fine-tune all the transformer encoder layers at a small learning rate and keep the bottom linear layer. About the \emph{Learning rate} shown in \ref{fig:lr}, the best learning rate is 1e-7, which can not be too large or too small.A large learning rate hurts the performance. Even more, it can not leverage the advantage of pre-trained weights.
	\begin{table}[!t]
		\setlength{\tabcolsep}{5pt}
		\centering
		\scalebox{0.75}{
		\begin{tabular}{ccccccc}
			\toprule
			Pretrain & P-PT & R@1$\uparrow$   & R@5$\uparrow$ & R@10$\uparrow$ & MdR$\downarrow$ & MnR$\downarrow$  \\ \midrule
			\multicolumn{7}{c}{MSR-VTT} \\
			ZS & & 30.6 & 54.4 & 64.3 & 4 & 41.8 \\
			ZS &\checkmark & 32.0 & 57.0 & 66.9 & 4 & 34.0 \\
			FT & & 43.1 & 70.4 & \textbf{80.8} & \textbf{2} & \textbf{16.2} \\
			FT &\checkmark&  \textbf{43.5} & \textbf{70.7} & 80.5 & \textbf{2} & 16.3 \\
			\midrule
			\multicolumn{7}{c}{MSVD} \\
			ZS & & 36.2 & 63.8 & 73.5 & 3 & 20.4 \\
			ZS &\checkmark & 38.5 & 66.9 & 76.8 & 2 & 17.8 \\
			FT && 46.2 & \textbf{76.1} & 84.6 & \textbf{2} & 10.0 \\
			FT &\checkmark& \textbf{46.6} & \textbf{76.1} & \textbf{84.8} & \textbf{2} & \textbf{9.9} \\
			\midrule
			\multicolumn{7}{c}{LSMDC} \\
			ZS & & 13.6 & 27.9 & 35.5 & 32 & 134.5 \\
			ZS &\checkmark & 15.1 & 28.5 & 36.4 & 28 & 117.0 \\
			FT & & 20.7 & 38.9 & 47.2 & 13 & 65.3 \\
			FT &\checkmark& \textbf{21.7} & \textbf{39.5} & \textbf{49.1} & \textbf{11} & \textbf{61.2} \\
			\bottomrule
		\end{tabular}
		}
		\caption{Test on post-pretraining (P-PT) on (Ours)-meanP model with HowTo100M-380k dataset. ZS: zero-shot, FT: fine-tuning.}
		\label{tab:study_of_retrieval_pretrain_MSR-VTT}
	\end{table}
	\begin{table}[!t]
		\setlength{\tabcolsep}{5pt}
		\centering
		\scalebox{0.85}{
		\begin{tabular}{cccccc}
			\toprule
			2D/3D & R@1$\uparrow$   & R@5$\uparrow$ & R@10$\uparrow$ & MdR$\downarrow$ & MnR$\downarrow$  \\ \midrule
			\multicolumn{6}{c}{MSR-VTT} \\
			2D & 43.1 & 70.4 & 80.8 & 2 & 16.2 \\
			3D & 41.6 & 69.9 & 79.5 & 2 & 17.3 \\
			\midrule
			\multicolumn{6}{c}{MSVD} \\
			2D & 46.2 & 76.1 & 84.6 & 2 & 10.0 \\
			3D & 44.0 & 73.6 & 83.0 & 2 & 11.3 \\
			\midrule
			\multicolumn{6}{c}{LSMDC} \\
			2D & 20.7 & 38.9 & 47.2 & 13 & 65.3 \\
			3D & 20.8 & 40.6 & 49.3 & 11 & 61.0 \\
			\bottomrule
		\end{tabular}
		}
		\caption{Test 2D and 3D patch linear on (Ours)-meanP.}
		\label{tab:study_of_retrieval_3d}
	\end{table}

	\subsection{Post-pretraining on Video Dataset}
	Our model is built on the pre-trained CLIP, which is an image pre-training model. To solve this data type (image v.s. video) variance, we conduct a preliminary exploration on the post-pretraining of our model on Howto100M-380k video dataset, and report the results for both zero-shot and fine-tune. From the table \ref{tab:study_of_retrieval_pretrain_MSR-VTT}, we can see that the performance increases for both zero-shot and fine-tuning setting. The increase of zero-shot is much larger, which shows that post-pretraining with the same data type (video) can learn general knowledge and directly transfer to the task. In addition, the fine-tuning on the post-pretrained model also improves the performance on both LSMDC and MSVD datasets and achieves approximate results for the MSR-VTT dataset. In future work, we will explore the capability of pre-training with an even larger dataset.

	\subsection{Sampling strategy}
	We conduct three different sampling strategies for video. `Head' is to sample the first frames at the beginning of the video, `Tail' is to select the last frames at the end of the video, and `Uniform' is to sample the whole frames of the video uniformly. The experimental results show that `Uniform' is relatively a good choice, and `Head' is comparable. The `Tail' sampling strategy is unlikely to be used.

	\subsection{2D/3D Patch Linear}
	We conduct the comparison on 2D and 3D linear mentioned in Section \ref{sec:video_encoder}. Table \ref{tab:study_of_retrieval_3d} presents the performance of them. Against our expectation that 3D patch linear can extract temporal information among frames and generate better discriminant features and performance, the 3D linear generate worse results than 2D linear on both MSR-VTT and MSVD. We suppose that the CLIP is trained for 2D linear instead of 3D linear, and the discrepant initialization on 3D linear makes it hard to learn the temporal information. We will pretrain on a large video-text dataset to unleash its potential in future work.

	\section{Conclusion}
	In this paper, we use the pretrained CLIP as our backbone to solve the video clip retrieval task from frame-level input. We employ parameter-free type, sequential type, and tight type similarity calculator to obtain the final results. The experimental results demonstrate the effectiveness of our model and achieve the SOTA results on MSR-VTT, MSVC, LSMDC, ActivityNet, and DiDeMo. Besides, we give serval insights from our empirical studies: 1) image feature can also promote the video-text retrieval, 2) post-pretrain on even outstanding image-text pretrained CLIP can further improve the performance on video-text retrieval, 3) 3D patch linear projection and sequential type similarity are promising approaches on the retrieval task, and 4) The CLIP used on video-text retrieval is learning-rate sensitivity.
	
	% Entries for the entire Anthology, followed by custom entries
	\bibliography{End2EndVideoRetrieval}
	\bibliographystyle{acl_natbib}

	\clearpage
	\appendix
	\setcounter{table}{0}
	\renewcommand{\thetable}{A\arabic{table}}
	\section{Video-to-Text Retrieval}
	Table \ref{tab:video_to_text_retrieval_MSR-VTT}-\ref{tab:video_to_text_retrieval_lsmdc} present the video-to-text retrieval results of CLIP4Clip on MSR-VTT, LSMDC, MSVD, ActivityNet, and DiDeMo.
	\begin{table}[!htp]
		\setlength{\tabcolsep}{2pt}
		\centering
		\scalebox{0.70}{
			\begin{tabular}{lccccccc}
				\toprule 
				Methods        & TrainD & E2E & R@1$\uparrow$   & R@5$\uparrow$ & R@10$\uparrow$ & MdR$\downarrow$ & MnR$\downarrow$  \\ \midrule
				\multicolumn{8}{c}{Zero-shot} \\
				CLIP-straight$^a$		& W & \checkmark & 27.2 & 51.7 & 62.6 & 5 & -  \\
				\midrule
				\multicolumn{8}{c}{Training-7K} \\
				HowTo100M$^b$			& H+M & \checkmark & 16.8 & 41.7 & 55.1 & 8 & -  \\
				\midrule
				\multicolumn{8}{c}{Training-9K} \\
				CE$^c$					& M & & 20.6 & 50.3 & 64.0 & 5.3 & - \\ 
				MMT$^d$					& H+M & & 27.0 & 57.5 & 69.7 & 3.7 & -  \\
				AVLnet$^e$				& H+M & & 28.5 & 54.6 & 65.2 & 4 & -  \\
				SSB$^f$					& H+M & & 28.5 & 58.6 & 71.6 & 3 & -  \\
				HiT$^g$					& H+M & & 32.1 & 62.7 & 74.1 & 3 & - \\
				TT-CE+$^h$	& M & & 32.1 & 62.7 & 75.0 & 3 & - \\
				\cdashline{1-8}[3pt/4pt]
				(Ours)-meanP					& W+M & \checkmark & 43.1 & 70.5 & 81.2 & 2 & 12.4 \\
				(Ours)-seqLSTM					& W+M & \checkmark & 42.8 & 71.0 & 80.4 & 2 & 12.3 \\
				(Ours)-seqTransf				& W+M & \checkmark & 42.7 & 70.9 & 80.6 & 2 & 11.6 \\
				(Ours)-tightTransf				& W+M & \checkmark & 40.6 & 69.5 & 79.5 & 2 & 13.6 \\
				\bottomrule
			\end{tabular}
		}
		\caption{Results of video-to-text retrieval on MSR-VTT dataset. `Training-7K' follows the data splits from \cite{miech2019howto100m} and `Training-9K' follows the data splits from \cite{Gabeur2020MMT}. They have the same test set but different training set. The column `TrainD' shows the datasets used for pre-training and training, where M, H, W denote MSR-VTT, HowTo100M \cite{miech2019howto100m} and WIT \cite{radford2021learning}. The column `E2E' with \checkmark means training from raw video in an end-to-end manner. The baseline methods are $^a$CLIP-straight \cite{PortilloQuintero2021}, $^b$HowTo100M \cite{miech2019howto100m}, $^c$CE \cite{Liu2019CE}, $^d$MMT \cite{Gabeur2020MMT}, $^e$AVLnet \cite{Rouditchenko2020}, $^f$SSB \cite{patrick2021supportset},  $^g$HiT \cite{Liu2021HiT}, $^h$TT-CE+ \cite{croitoru2021teachtext}.}
		\label{tab:video_to_text_retrieval_MSR-VTT}
	\end{table}
	\begin{table}[!htp]
		\setlength{\tabcolsep}{2pt}
		\centering
		\scalebox{0.70}{
		\begin{tabular}{lccccccc}
			\toprule
			Methods  & TrainD & E2E & R@1$\uparrow$   & R@5$\uparrow$ & R@10$\uparrow$ & MdR$\downarrow$ & MnR$\downarrow$  \\ \midrule
			CLIP-straight$^a$   & W & \checkmark & 59.9 & 85.2 & 90.7 & 1 & -  \\
			TT-CE+$^b$	& W & & 27.1 & 55.3 & 67.1 & 4 & - \\
			\midrule
			(Ours)-meanP					& W+M & \checkmark & 56.6 & 79.7 & 84.3 & 1 & 7.6 \\
			(Ours)-seqLSTM					& W+M & \checkmark & 52.5 & 74.0 & 78.1 & 1 & 14.7 \\
			(Ours)-seqTransf				& W+M & \checkmark & 62.0 & 87.3 & 92.6 & 1 & 4.3 \\
			(Ours)-tightTransf				& W+M & \checkmark & 54.3 & 85.3 & 91.0 & 1 & 6.0 \\
			\bottomrule
		\end{tabular}
		}
		\caption{Results of video-to-text retrieval on MSVD dataset. In the column `TrainD', M and W denote training on MSVD and WIT \cite{radford2021learning}, and CW means CC3M \cite{sharma2018conceptual} plus WebVid-2M \cite{Bain2021Frozen}. The column `E2E' with \checkmark means training from raw video in an end-to-end manner. The baseline method is $^a$CLIP-straight \cite{PortilloQuintero2021}, $^b$TT-CE+ \cite{croitoru2021teachtext}.}
		\label{tab:video_to_text_retrieval_msvd}
	\end{table}
	\newpage
	\begin{table}[!htp]
		\setlength{\tabcolsep}{2pt}
		\centering
		\scalebox{0.70}{
		\begin{tabular}{lccccccc}
			\toprule
			Methods  & TrainD & E2E & R@1$\uparrow$   & R@5$\uparrow$ & R@10$\uparrow$ & MdR$\downarrow$ & MnR$\downarrow$  \\ \midrule
			JSFusion$^a$  & L & \checkmark & 12.3 & 28.6 & 38.9 & 20 & - \\
			CLIP-straight$^b$   & L & \checkmark & 6.8 & 16.4 & 22.1 & 73 & - \\
			TT-CE+$^c$	& L & & 17.5 & 36.0 & 45.0 & 14.3 & - \\
			\midrule
			(Ours)-meanP					& W+L & \checkmark & 20.6 & 39.4 & 47.5 & 13 & 56.7 \\
			(Ours)-seqLSTM					& W+L & \checkmark & 20.9 & 40.7 & 49.1 & 11 & 53.9 \\
			(Ours)-seqTransf				& W+L & \checkmark & 20.8 & 39.0 & 48.6 & 12 & 54.2 \\
			(Ours)-tightTransf				& W+L & \checkmark & 17.4 & 36.7 & 45.0 & 15 & 65.3 \\
			\bottomrule
		\end{tabular}
		}
		\caption{Results of video-to-text retrieval on LSMDC dataset. In the column `TrainD', L and W denote training on LSMDC and WIT \cite{radford2021learning}, MD used in \cite{Dzabraev2021MDMMT} denotes a combined multidomain dataset containing MSR-VTT, LSMDC, HowTo100M, etc., and CW means CC3M \cite{sharma2018conceptual} plus WebVid-2M \cite{Bain2021Frozen}. The column `E2E' with \checkmark means training from raw video in an end-to-end manner. The baseline methods are $^a$JSFusion \cite{yu2018joint}, $^b$CLIP-straight \cite{PortilloQuintero2021}, $^c$TT-CE+ \cite{croitoru2021teachtext}.}
		\label{tab:video_to_text_retrieval_lsmdc}
	\end{table}
	\begin{table}[!htp]
		\setlength{\tabcolsep}{2pt}
		\centering
		\scalebox{0.70}{
			\begin{tabular}{lccccccc}
				\toprule
				Methods  & TrainD & E2E & R@1$\uparrow$   & R@5$\uparrow$ & R@10$\uparrow$ & MdR$\downarrow$ & MnR$\downarrow$  \\ \midrule
				FSE$^a$	 & A & & 16.7 & 43.1 & - & 7.0 & - \\ 
				CE$^b$   & A & & 17.7 & 46.6 & - & 6.0 & 24.4\\ 
				HSE$^a$  & A & & 18.7 & 48.1 & - & -  & -\\ 
				MMT$^c$  & H+A & & 28.9 & 61.1 & - & 4.0 & 17.1\\ 
				SSB$^d$  & H+A & & 28.7 & 60.8 & - & 2.0 & -\\  
				TT-CE+$^e$	& A & & 23.0 & 56.1 & - & 4.0 & - \\
				\midrule
				(Ours)-meanP					& W+A & \checkmark & 42.5 & 74.1 & 85.8 & 2.0 & 6.6 \\
				(Ours)-seqLSTM					& W+A & \checkmark & 42.6 & 73.4 & 85.6 & 2.0 & 6.7 \\
				(Ours)-seqTransf				& W+A & \checkmark & 41.4 & 73.7 & 85.3 & 2.0 & 6.7 \\
				(Ours)-tightTransf				& W+A & \checkmark & 18.9 & 49.6 & 65.8 & 6.0 & 16.3 \\
				\bottomrule
			\end{tabular}
		}
		\caption{Results of video-to-text retrieval on ActivityNet dataset. In the column `TrainD', A, H, and W denote training on ActivityNet, HowTo100M \cite{miech2019howto100m}, and WIT \cite{radford2021learning}. The column `E2E' with \checkmark means training from raw video in an end-to-end manner. The baseline methods are $^a$FSE,HSE \cite{bowen2018cross}, $^b$CE \cite{Liu2019CE}, $^c$MMT \cite{Gabeur2020MMT}, $^d$SSB \cite{patrick2021supportset}, $^e$TT-CE+ \cite{croitoru2021teachtext}.}
		\label{tab:video_to_text_retrieval_activityNet}
	\end{table}
	\newpage
	\begin{table}[!htp]
		\setlength{\tabcolsep}{2pt}
		\centering
		\scalebox{0.70}{
			\begin{tabular}{lccccccc}
				\toprule
				Methods  & TrainD & E2E & R@1$\uparrow$   & R@5$\uparrow$ & R@10$\uparrow$ & MdR$\downarrow$ & MnR$\downarrow$  \\ \midrule
				S2VT$^a$      & D & \checkmark & 13.2 & 33.6 & - & 15.0 & - \\ 
				FSE$^b$	      & D & & 13.1 & 33.9 & - & 12.0 & - \\ 
				CE$^c$        & D & & 15.6 & 40.9 & - & 8.2  & 42.4\\ 
				TT-CE+$^d$	  & D & & 21.1 & 47.3 & 61.1 & 6.3 & - \\
				\midrule
				(Ours)-meanP					& W+D & \checkmark & 42.5 & 70.6 & 80.2 & 2.0 & 11.6 \\
				(Ours)-seqLSTM					& W+D & \checkmark & 42.4 & 69.2 & 79.2 & 2.0 & 11.8 \\
				(Ours)-seqTransf				& W+D & \checkmark & 41.4 & 68.2 & 79.1 & 2.0 & 12.4 \\
				(Ours)-tightTransf				& W+D & \checkmark & 21.5 & 51.1 & 64.8 & 5.0 & 22.4 \\
				\bottomrule
			\end{tabular}
		}
		\caption{Results of video-to-text retrieval on DiDeMo dataset. In the column `TrainD', D and W denote training on DiDeMo and WIT \cite{radford2021learning}. The column `E2E' with \checkmark means training from raw video in an end-to-end manner. $\dagger$ means that the candidate video is concatenated using ground truth proposals. The baseline methods are $^a$S2VT \cite{Venugopalan2015Translating}, $^b$FSE \cite{bowen2018cross}, $^c$CE \cite{Liu2019CE}, $^d$ClipBERT \cite{lei2021less}, $^e$Frozen \cite{Bain2021Frozen}, $^d$TT-CE+ \cite{croitoru2021teachtext}.}
		\label{tab:video_to_text_retrieval_didemo}
	\end{table}

\end{document}